\def\cvprPaperID{6138}
\def\confName{ICCV}
\def\confYear{2023}

\def\paperTitle{LivelySpeaker: Towards Semantic-Aware Co-Speech Gesture Generation}

\def\authorBlock{
    Yihao Zhi\textsuperscript{\rm 1,}\thanks{Equal contribution} \quad
    Xiaodong Cun\textsuperscript{\rm 2,}\footnotemark[1] \quad
    Xuelin Chen\textsuperscript{\rm 2} \quad
    Xi Shen\textsuperscript{\rm 3} \quad \\
    Wen Guo\textsuperscript{\rm 4} \quad
    Shaoli Huang\textsuperscript{\rm 2} \quad
    Shenghua Gao\textsuperscript{\rm 1,5,6,}\thanks{Corresponding author.} \\ \\
    \textsuperscript{\rm 1}ShanghaiTech University \quad
    \textsuperscript{\rm 2}Tencent AI Lab \quad
    \textsuperscript{\rm 3}Intellindust \quad
    \textsuperscript{\rm 4}INRIA \quad \\
    \textsuperscript{\rm 5}Shanghai Engineering Research Center of Intelligent Vision and Imaging \\
	\textsuperscript{\rm 6}Shanghai Engineering Research Center of Energy Efficient and Custom AI IC\\
    \\
    \url{https://github.com/zyhbili/LivelySpeaker}
}

\newif\ifreview 
\newif\ifarxiv \newcommand{\arxiv}{\arxivtrue}
\newif\ifcamera 
\newif\ifrebuttal

\newcommand{\xl}[1]{{\color{black}#1}}

\arxiv 
\pdfoutput=1
\documentclass[10pt,twocolumn,letterpaper]{article}
\ifreview \usepackage[review]{cvpr} \fi
\ifarxiv \usepackage[pagenumbers]{cvpr} \fi
\ifrebuttal \usepackage[rebuttal]{cvpr} \fi
\ifcamera \usepackage{cvpr} \fi

\usepackage{graphicx}
\usepackage{amsmath}
\usepackage{amssymb}
\usepackage{booktabs}

\usepackage{times}
\usepackage{microtype}
\usepackage{epsfig}
\usepackage[table,xcdraw]{xcolor}
\usepackage{caption}
\usepackage{float}
\usepackage{placeins}
\usepackage{color, colortbl}
\usepackage{stfloats}
\usepackage{enumitem}
\usepackage{tabularx}
\usepackage{xstring}
\usepackage{multirow}
\usepackage{xspace}
\usepackage{url}
\usepackage{subcaption}
\usepackage{xcolor}

\usepackage{amsmath}

\usepackage[hang,flushmargin]{footmisc}

\ifcamera \usepackage[accsupp]{axessibility} \fi

% ...

\ifarxiv  \fi

\newcommand{\R}[1]{{%
    \textbf{%
        \ifstrequal{#1}{1}{\textcolor[rgb]{1,0.33,0.2}{R#1}}{%
        \ifstrequal{#1}{2}{\textcolor[rgb]{0.2,0.7,0.1}{R#1}}{%
        \ifstrequal{#1}{3}{\textcolor{cyan}{R#1}}{%
        \ifstrequal{#1}{4}{\textcolor{teal}{R#1}}{%
                           \textcolor{cyan}{R#1}%
        }}}}%
    }%
}}

\usepackage{xr-hyper}

\makeatletter
\newcommand*{\addFileDependency}[1]{
  \typeout{(#1)}
  \@addtofilelist{#1}
  \IfFileExists{#1}{}{\typeout{No file #1.}}
}

\makeatother

\usepackage[pagebackref,breaklinks,colorlinks]{hyperref}
\usepackage[capitalize]{cleveref}
\crefname{section}{Sec.}{Secs.}
\crefname{table}{Table.}{Tables}
\crefname{figure}{Fig.}{Figs.}

\hypersetup{
    colorlinks = true,
    linkbordercolor = {white},
    citecolor=blue,
    urlcolor=magenta
}

\begin{document}
\title{\paperTitle}
\author{\authorBlock}
\twocolumn[{
\maketitle
\begin{center}
    \captionsetup{type=figure}
    \includegraphics[width=1.\linewidth]{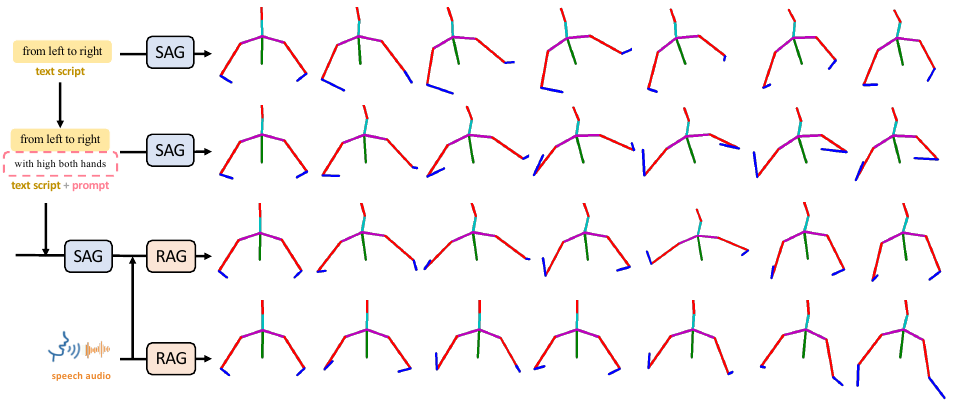}
    \vspace{-3.em}
    \captionof{figure}{
    We propose LivelySpeaker, a novel system that decouples the co-speech gesture generation into two stages, namely semantic-aware generator~(SAG) and rhythm-aware generator~(RAG), respectively. 
    Powered by the proposed two-stage framework, our method can generate semantic-aware gestures~(top three rows) other than purely audio-driven results~(bottom row). 
    We can also add additional prompts to the text script to specify the gestures~(second line). 
    Last, when we integrate the proposed components altogether, our method can still retain the semantic gestures and also appropriate rhythm for a lively speaker~(third row).
    }
    \label{fig:teaser}
\end{center}
}]

\renewcommand{\thefootnote}{\fnsymbol{footnote}}
\footnotetext[1]{~Equal contribution.}
\footnotetext[2]{~Corresponding author.}

\begin{abstract}

Gestures are non-verbal but important behaviors accompanying people's speech. While previous methods are able to generate speech rhythm-synchronized gestures, the semantic context of the speech is generally lacking in the gesticulations. Although semantic gestures do not occur very regularly in human speech,
they are indeed the key for the audience to understand the speech context in a more immersive environment. Hence, we introduce LivelySpeaker, a framework that realizes semantics-aware co-speech gesture generation and offers several control handles. In particular, our method decouples the task into two stages: script-based gesture generation and audio-guided rhythm refinement. Specifically,
the script-based gesture generation leverages the pre-trained CLIP text embeddings as the guidance for generating gestures that are highly semantically aligned with the script. Then, we devise a simple but effective diffusion-based gesture generation backbone simply using pure MLPs, that is conditioned on only audio signals and learns to gesticulate with realistic motions. We utilize such powerful prior to rhyme the script-guided gestures with the audio signals, notably in a zero-shot setting. Our novel two-stage generation framework also enables several applications, such as changing the gesticulation style, editing the co-speech gestures via textual prompting, and controlling the semantic awareness and rhythm alignment with guided diffusion. Extensive experiments demonstrate the advantages of the proposed framework over competing methods.
In addition, our core diffusion-based generative model also achieves state-of-the-art performance on two benchmarks. The code and model will be released to facilitate future research.
\end{abstract}

\section{Introduction}
During human conversation, non-verbal behaviors are typically present and among them, the most significant is gesture language.
These non-linguistic gestures serve as an auxiliary but effective means of conveying key messages, enriching the conversation with contextual cues, and facilitating better understanding among participants.~\cite{goldin2013gesture,cassell1999speech, hostetter2008visible, iverson1998people}.
Empowering the digital replicas of humans with the ability to gesticulate has been a long pursuit in the research community, as such ability can benefit many downstream applications, including digital humans in the coming virtual universe, non-player game characters, robot assistants, \etc.

Given the speech content in the form of texts and/or audio streams, the objective is to generate realistic co-speech gestures.
Traditional methods achieve this with hard-coded rules~\cite{10.1145/192161.192272, Cassell2004,huang2012robot, 10.1145/2485895.2485900},
\eg, ``\textit{good}'' in the speech will be \xl{simply} represented by the gesture ``\textit{thump up}''.
However, these methods usually produce deterministic results; more importantly, they can not guarantee smooth transitions in the results.
Recently, deep learning-based methods have been prevalent in the field of gesture generation from multi-modality inputs.
In particular, these methods formulate the problem as conditional motion generation and tackle it via training a conditional generative model that takes as input the speaker identities~\cite{speech2gesture}, audio waves~\cite{sdt}, speech texts~\cite{text2gestures}, or a combination of these multi-modal signals~\cite{ha2g, trimodal, ao2022rhythmic}.
Although multiple modalities are incorporated in the formulation, the results are often dominated by the rhythm of the audio signal since it is highly correlated with the performance of gestures during speech. While other works recognize the importance of the semantics conveyed through co-speech gestures, their framework heavily depends on pre-defined gesture types~\cite{liang2022seeg, bhattacharya2021speech2affectivegestures} or keywords~\cite{xu2022freeform}, making it difficult to express more complex intentions effectively.

\begin{figure}
    \centering
    \includegraphics[width=\linewidth]{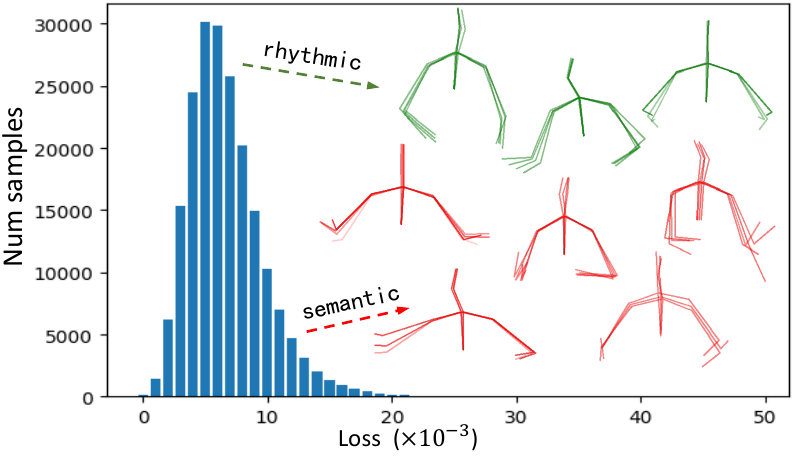}
    \vspace{-2em}
    \caption{We plot the $L_2$ loss histogram on the training samples of the pre-trained trimodal method~\cite{trimodal}. Although it is learned from multiple conditions~(\eg, text, audio), their methods are still dominated by repeated rhythm, and hard to model the rarely appeared diversity gestures, \eg, the semantic-aware motions.}
    \label{fig:distribution}
\end{figure}

We begin with insights from the following two perspectives:
Real-world human conversations contain a limited number of semantic gestures (See ~\cref{fig:distribution}), which presents difficulties in learning co-speech gestures that are semantic-sensitive but rhythm-irrelevant. This partially explains why prior approaches have yielded results that heavily rely on the audio rhythm.
$(ii)$ Most previous methods are built on generative adversarial networks~(GANs), which might be hard to train, especially when learning a many-to-many mapping between the text/audio and the gesture~\cite{sdt}. 

Following this, we present LivelySpeaker, a simple and effective framework for semantic-aware co-speech gesture generation.
In particular,
our framework \emph{explicitly} decouples the generation into two stages, namely the \emph{script-based gesture generation}, and the \emph{audio-guided rhythm refinement}. 
Specifically,
the script-based gesture generation leverages the pre-trained CLIP~\cite{radford2021learning} text embeddings as the guidance for generating gestures that are highly semantically correlated with the textual script.
In the second stage, we devise a simple but effective diffusion-based gesture generation backbone with pure MLPs, that is conditioned on only audio signals and learns to gesticulate with realistic motions.
We utilize such powerful prior to rhyme the script-guided gestures with the audio signals, notably in a zero-shot setting.
In detail,
we gradually add Gaussian noise for $T$ steps to the motion extracted from the dataset,
on which an MLP-based~\cite{guo2023back} motion denoising model~\cite{tevet2022MDM} is conditioned on the corresponding audio and predicts the clean motion.
We show that this diffusion-based model is effective in rhyming rather smooth gestures produced from the script-based generation module with the audio signals.

Building upon these two powerful modules, our method can generate diverse and high-quality co-speech gestures that are semantically meaningful, given the textual description of the speech and audio.
Extensive experiments show that the proposed framework yields state-of-the-art performance in co-speech gesture generation.
We also conduct experiments to show the control ability of our method by extending it to a number of scenarios that are not possible with competing methods, 
including changing the gesticulation style, editing the co-speech gestures via textual prompting, and controlling the semantic awareness and rhythm alignment with guided diffusion.

The main contribution of this paper is summarised: 
\begin{itemize}
    \item We propose LivelySpeaker, a novel two-stage framework for semantic-aware and rhythm-aware co-speech gesture generation.
    \item A novel MLP-based diffusion-based backbone is devised, that achieves state-of-the-art performance on the two benchmarks for co-speech generation.
    \item Our framework enables several new applications in co-speech gesture generation, such as text prompt-based gesture control, balancing the control between two different condition modalities (\ie, text and audio).
\end{itemize}

\section{Related work}
\noindent\textbf{Co-speech Gesture Generation.}
As aforementioned, 
the research of co-speech gesture generation has taken several routes in the past decades, including rule-based \cite{prendinger2004mpml, kipp2005gesture, kopp2006towards}, machine learning-based \cite{huang2014learning, sargin2008analysis, levine2010gesture}, and deep learning-based ones \cite{bhattacharya2021speech2affectivegestures, audio2gestures, speech2gesture, sdt, trimodal, ha2g}.
In our coverage, we mainly review the deep learning-based ones, as they have shown better performance and are more relevant to our method.
Earlier works along this line consider the problem as an end-to-end regression of 2D keypoints of the human body, with different dedicated designs on the network architectures.
For example, Speech2Gesture~\cite{speech2gesture} generates personalized 2D keypoints from audio using a conditional generative adversarial network.
Ahuja \etal \cite{ahuja2020style} propose a few-shot method for personalized motion transfer.
Furthermore, being aware of the many-to-many nature of the co-speech gesture generation, \cite{sdt} proposes a novel audio template-based method to reduce the uncertainty of the generation.
Going beyond simple 2D keypoints, \cite{liu2022audio} proposes a co-speech generation framework that leverages an unsupervised motion representation instead of a structural human body prior and involves the image-based rendering technique to generate co-speech videos like talking face generation~\cite{sadtalker}. 
These methods are limited in their applicability to many real-world scenarios as they generate motions of 2D key points or directly output 2D imagery of the speaker.

To generate motions for 3D avatars,
TriModal~\cite{trimodal} extracts diverse upper body motions from the TED talks and designs an LSTM-based neural network that is conditioned on the audio, text, and identity and generates co-speech gestures.
Speech2AffectiveGesture~\cite{bhattacharya2021speech2affectivegestures} extends this work for more semantic-aware gesture generation. Nevertheless, they have only demonstrated on five predefined gesture classes. 
As a follow-up that increases the effectiveness of Speech2AffectiveGesture, HA2G~\cite{ha2g} further extracts the hand keypoints on TED datasets and uses a hierarchical GRU~\cite{cho2014properties} network.
More recently, Ao \etal~\cite{ao2022rhythmic} propose a method using VQ-VAE~\cite{vqvae},
and SEEG~\cite{liang2022seeg} is designed to generate semantic gestures of several kinds. 
In summary,
most of these methods learn in an end-to-end fashion, where the conditional audio in fact dominates the conditional generation, and only focuses on limited types of semantic gestures~\cite{bhattacharya2021speech2affectivegestures,liang2022seeg}.
Differently, our framework can produce co-speech gestures that are highly semantically aligned with the textual description of the speech provided by the user.

\noindent\textbf{Conditional Motion Generation.}
Co-speech generation is also a sub-topic of human motion generation, which aims at generating 3D human motion from various conditions. 
One hottest topic is text-to-motion. 
Language2Pose~\cite{ahuja2019language2pose} employs a curriculum learning approach to learn a joint embedding space for both text and pose. The decoder can thus take text embedding to generate motion sequences. Ghost \textit{et al.}~\cite{ghosh2021synthesis} extend it through manifold representations for the upper body and the lower body movements. Similarly, MotionCLIP~\cite{tevet2022motionclip} also tends to align text and motion embedding but proposes to utilize CLIP~\cite{radford2021learning} as the text encoder and employ rendered images as extra supervision. It shows the ability to generate out-of-distribution motion and enable latent code editing. 
As for the generation of the simple action, several methods~\cite{petrovich21actor, petrovich22temos,TEACH:3DV:2022, t2mgpt} have been proposed by a pre-defined action class~\cite{petrovich21actor}, an additional text encoder~\cite{petrovich22temos} and temporal motion compositions from a series of natural language descriptions~\cite{TEACH:3DV:2022}. Guo \textit{et al.}~\cite{guo2022generating} also proposes to incorporate motion length prediction from text to produce motion with reasonable length.  
Motion can also be generated with music ~\cite{lee2019dancing,li2020learning,li2021ai,aristidou2021rhythm,chen2021choreomaster,siyao2022bailando}, which has a similar form as our task but they only focus on the rhythm. \eg, \cite{chen2021choreomaster} produce a graph-based network to optimize the choreography-aware features,  Li \etal~\cite{li2021ai} generate the dance motions from a transformer-based network and a high-quality dataset. Li \etal~\cite{siyao2022bailando} train a separate VQ-VAE~\cite{vqvae} to model the upper body and lower body individually.
Through sharing a similar goal with these works, our work differs from them as we consider both the semantic and rhythm awareness in our unified framework.

\noindent\textbf{Diffusion-based Motion Generation.}
Very recently, the denoising diffusion-based models have also shown very promising results in generating human motion. Prior works of MDM~\cite{tevet2022MDM} and MotionDiffuse~\cite{zhang2022motiondiffuse} generate realistic motions from noise inspired by the denoising diffusion model~\cite{ddpm}. PhysDiff~\cite{physdiff} extend MDM via the physics-aware restriction. EDGE~\cite{edge} design a stronger dance generation network using the powerful pre-trained audio model, jukebox~\cite{jukebox}. These methods only use the diffusion-based model for conditional generation, whereas our method finds more interesting applications via the diffusion-based model in motion synthesis.

\section{Method}
\begin{figure*}
    \centering
    \includegraphics[width=\linewidth ]{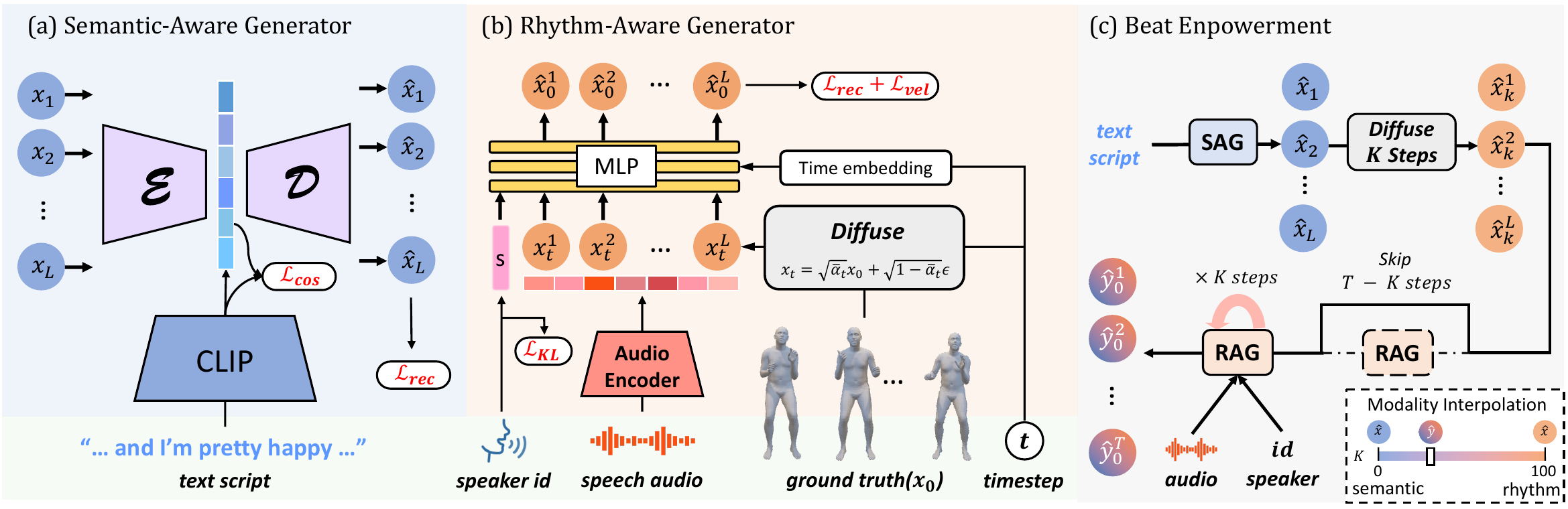}
    \caption{We propose a two-stage framework for semantic-aware gesture generation~(a) and rhythm-aware gesture generation~(b), respectively. Then, we combine them via a beat empowerment method as in~(c) for the full pipeline of the proposed LivelySpeaker.}
    \label{fig:overview}
\end{figure*}

Given the speech content in the form of audio, its corresponding text script, and the identity information, our system aims to generate 3D skeletal gestures that are semantically and rhythmically aligned with the speech. 

We tackle this problem with a two-stage framework consisting of a semantic-aware generator (SAG), and a rhythm-aware generator (RAG), as shown in Fig.~\ref{fig:overview}~(a) and (b), respectively. 
After training each component, we can generate the gestures from the text scripts first, and then leverage the rhythm-aware network as a beat empowerer as in Fig.~\ref{fig:overview}~(c). 
In the following, we first present the details of the semantic-aware and rhythm-aware generator in \cref{sec3:sem} and \cref{sec3:rhy}, followed by details of the inference pipeline and application of the whole framework in \cref{sec3:app}.

\subsection{Generating Semantic Gestures from Text Script}
\label{sec3:sem}

Current co-speech generation approaches~\cite{ha2g,trimodal,liang2022seeg} often consider the script and audio features equally with identical timestamps concatenation.
In these approaches, the text features aligned with timestamps are more likely to act as alternative beat signals generating the rhythm-dominant gestures, which have a small effect on previous methods. 
To utilize the semantic information well, in the first stage of our framework, we only train a semantic-aware generator~(SAG) to generate the gesture from text scripts.

Inspired by the progress of text to motion~\cite{tevet2022motionclip,t2mgpt,tevet2022MDM}, we consider the text script as a kind of semantic description to generate the corresponding motion.
As shown in Fig.~\ref{fig:overview}~(a), we split the motion sequences into fixed segments and send them into an encoder-decoder-like Transformer~\cite{vaswani2017attention} for motion generation~\cite{bhattacharya2021speech2affectivegestures,trimodal}. Our network contains 3-layer encoders and decoders. Each Transformer layer has a latent dimension of 512, and the dimension of the feed-forward layer equals 1024. To integrate the semantic-aware information, we use a pre-trained CLIP~\cite{clip} of ViT-B/32 as the text embedding network, getting 512-dimension semantic features of the whole script sequences, other than the frame-wise semantic feature as in previous works~\cite{trimodal,ha2g}.

For training, we feed the ground truth pose sequences $x_{1:t}$  to the transformer encoder to generate the motion latent: $z_{emb}=E(x_{1:t}) \in R^{512}$, and a decoder is used to decode this latent code to reconstruct a sequence of poses $\hat{x}_{1:t}=D(E(x_{1:t}))$. Then, we calculate the distance between the semantic embeddings of CLIP $z_{CLIP}$ and the latent code $z_{emb}$ using a cosine similarity loss $L_{cos}$. We also measure the reconstruction loss $L_{rec}$ between the generated motion and the original one using simple mean square error. The full training objective of SAG is:
\begin{equation}
    L_{full} = L_{rec}(x_0, \hat{x_0})+\lambda L_{cos}(z_{CLIP}, z_{emb}) ,
\end{equation}
where we set $\lambda = 1$ empirically.
In testing, we generate the motion sequences directly from the CLIP embedding.

\subsection{Diffusion-Based Rhythm-Aware Generator}
\label{sec3:rhy}

Although our SAG can produce some semantic-aware gestures, the out-sync gestures also restrict the realism of the generated motion. However, it is hard to align the temporal information of the generated motion only and keep other content unchanged. We take advantage of the diffusion-based model for its powerful ability in distribution modeling~\cite{ddpm,tevet2022MDM} and editing~\cite{sdedit,ddim}.

The denoising diffusion model is a Markov noising process, which first shows its potential in image generation~\cite{ddpm}. Following the human motion diffusion model~\cite{tevet2022MDM}, the input pose sequences can be defined as $\{x_t^{1:N}\}_{t=0}^T$, where $x_0^{1:N}$ is sampled from the data distribution and
\begin{equation}
    q(x_t^{1:N}|x_{t-1}^{1:N}) = N(\sqrt{\alpha_{t}}x_{t-1}^{1:N}, (1 - \alpha_t)I).
\end{equation}
Here, $\alpha_t \in (0,1)$ are constant numbers. When $\alpha_t$ is sufficiently small, we can make an approximation that $x_t^{1:N}$ follows a normal distribution with mean $0$ and variance $I$. Henceforth, we will refer to the complete sequence at noise step $t$ as $x_t$. In our task, we also follow previous work~\cite{ddpm} to predict the signal itself, \ie, $\hat{x}^0 = G(x_t, t, c) $, where $c$ is the conditional audio and $G$ is our denoising backbone.

As for the network structure, our audio encoder $C$ is simply constructed using four 1d-convolution blocks activated by leaky ReLU, where we feed in a raw audio waveform and generate a sequence of 256-channel feature vectors. We have also tested other similar audio encoders in previous work~\cite{faceformer, ha2g}, however, there is no observed performance gain.
As for the denoising diffusion network~$G$, different from the original MDM~\cite{tevet2022MDM}, we use $N$ layers MLP-based network~\cite{guo2023back} to construct, which generates better rhythm and generate more smooth results. In detail,
we first use a linear layer to project input data to a higher-dimension latent space. After applying a series of MLP blocks, a last linear layer is used to project the latent feature back to poses as output. 
Each MLP block is composed of one FC layer for temporal merging and one FC layer for spatial merging. For each MLP block, we use layer normalization (LN)~\cite{ba2016layer} as pre-normalization, SiLU~\cite{ramachandran2017searching} as activation, and apply skip-connections~\cite{resnet}. 
As for the additional conditions, we concatenate the audio feature to the sampled pose and add the time step embedding $t_{emb}$ to each MLP block. We also embed the speaker id into the vector and calculate the style embedding $s$ through reparameterization, where $s$ is concatenated along the temporal dimension. 

For training the denoising network, we split the long motion sequence to the specific length and calculate the reconstruction loss via the Huber loss $L_{huber}$ ~\cite{trimodal} of the diffusion model as:
\begin{equation}
L_{rec} = E_{x_0 \sim q(x_0|c), t \sim [1,T]}[L_{huber}(x_0, \hat{x_0})].
\end{equation}
Similarly, we add velocity loss as:
\begin{equation}
L_{vel} = E_{x_0 \sim q(x_0|c), t \sim [1,T]}[L_{huber}(\dot{x_0}, \dot{\hat{x_0}})].
\end{equation}

Besides, since human motion is subject-related~\cite{ha2g, trimodal}, the Kullback–Leibler divergence~\cite{kl} $L_{KL}$ is used to regularize the distribution of all speakers on the $s$ embeddings.

Overall, the loss function of training the rhythm-aware generator can be written as:
\begin{equation}
    L_{full} = L_{rec} + \lambda L_{KL} + \beta L_{vel},
\end{equation}
where $\lambda$ and $\beta$ equals to $1e^{-2}$ and $1$, respectively. We follow the previous works \cite{trimodal,ha2g} and set the threshold for the $L_{huber}$ to 0.1. To generate longer motions, we concatenate 4 previous frames to achieve visual continuity similar to previous works ~\cite{speech2gesture,  trimodal, ha2g}.

\subsection{Full LivelySpeaker Pipeline and Applications}
\label{sec3:app}
After training both the semantic-aware and rhythm-aware generator models, we can utilize the latter to address the rhythm issues with the output of the former. In detail, as shown in ~\cref{fig:overview}~(c), after generating the semantic-aware motions from the SAG, following SDEdit~\cite{sdedit}, we can invert the generated motion by adding $K$ steps noises, and then, we consider this motion as the generated motion of $T-K$~($K=20$ in our cases), denoising it to a new distribution via the guidance of the audio using DDIM ($T=100$). When inferring a long sequence, we repeat the procedure mentioned above for each motion clip (consisting of 34 frames) sequentially and then concatenate them together. Thanks to the power of the diffusion-based model, this simple beat empowerment step keeps both the diversities from the semantic-aware generator and hugely increases the rhythm's alignment.

Since we learn each stage individually and each stage models a different distribution, our methods enable some interesting applications. Below, we give brief introductions:

\noindent\textbf{Semantic motion generation via new text prompts.}
We find that the individual learned semantic-aware generator is also a good controllable gesture generator. We can add some new text prompts to the CLIP encoder of our semantic-aware generator, and our method also generates the corresponding motions. We give a brief visualization in ~\cref{fig:teaser} where the generated gestures are in the new pose. This phenomenon reveals that even if the semantic information rarely appears in the dataset, it is also learned by our network. We give more examples in the supplementary videos to show the effectiveness of the proposed methods.

\noindent\textbf{Interpolating poses between different modalities.}
Our method enables the applications of generating different gestures of the script-based motion and the rhythm-based refinements by controlling the denoising steps of the diffusion model (see \cref{fig:overview}~(c)). We give the comparison and details in the supplementary materials.

\section{Experiments}

\subsection{Datasets} 
We validate our pipeline using two datasets, including the TED Gesture dataset~\cite{trimodal} and BEAT dataset~\cite{liu2022beat}. The TED Gesture dataset \cite{trimodal} contains 1766 videos sourced from online TED speech videos and utilizes three modalities: audio, text, and speaker identity. The human pose is represented by direction vectors of 10 upper body joints.

Besides the body movements, clean finger movements are also essential for a lively speaker's delivery. Therefore, instead of using the noisy TED-Expressive~\cite{ha2g} which captures the figure motion by OpenPose, we evaluate the performance of the newly introduced high-quality dataset BEAT~\cite{liu2022beat} for its high fidelity on hand poses. BEAT~\cite{liu2022beat} is constructed using a commercial MOCAP system, including additional annotations for emotion and semantic modalities. It captures the rotation angles of joints that are invariant to body shape. During training, we convert its origin Euler angle to rot6d representation \cite{zhou2019continuity} to ensure better convergence. 

Following the previous settings~\cite{ha2g, liu2022beat}, both datasets are resampled into 15 fps and divided into fragments of 34 frames in length with overlapping clips.

\subsection{Evaluation Metrics} 
We adopt three main metrics to evaluate the generation quality, including Frechet Gesture Distance (FGD)~\cite{trimodal}, Beat Consistency Score~(BC~\cite{ha2g}), and Diversity~\cite{ha2g} as previous methods~\cite{trimodal, ha2g}. FGD is a metric to measure the distribution disparity between generated output and ground truth across the entire dataset, where a pre-trained autoencoder is used to project motion into latent space. On TED~\cite{trimodal} dataset, we use the autoencoder provided in \cite{trimodal} for a fair comparison, while on BEAT \cite{liu2022beat} we re-train the autoencoder using rot6d representation. Beat Consistency Score calculates the average distance between every audio beat and its nearest motion beat. Intuitively, the denser the motion beats are, the better BC. Thus, it would be invalid in cases of anomalous gesture sequences that contain numerous motion beats. Fortunately, we can take FGD as a reference to solve it. Diversity is assessed by measuring the variations in generated gestures, which are also calculated using a pre-trained autoencoder~\cite{trimodal}. It is computed by the $L_1$ distance between randomly sampled motion feature pairs \cite{ha2g}.

\subsection{implementation Details}
The training is composed of two stages. For Semantic Aware Generator, we train it with the Adam optimizer ($lr = 0.0001, \beta = (0.9, 0.99)$) for 400 epochs.
As for Rhythm Aware Generator, we train it with the AdamW optimizer ($lr = 0.0001,  \beta = (0.9, 0.999)$) for 1200 epochs. The total diffusion steps $T=1000$ in training, in inference, we generate the motion via 20 steps DDIM sampler. 
All experiments are conducted with a batch size of 512 on a single NVIDIA A100. When evaluating metrics, we use DDIM \cite{ddim} with 100 steps for faster sampling.

\subsection{Baselines}
We compare our method with the following methods. \textbf{Speech2Gestures}~\cite{speech2gesture} and \textbf{Trimodal}~\cite{trimodal} are two representative methods in co-speech gesture generation. Trimodal fuses three-modality information and achieves better performance than S2G. \textbf{HA2G}~\cite{ha2g}, the SOTA model on TED Gesture dataset, implements a coarse-to-fine hierarchical gesture generator and learns a powerful audio extractor through contrastive learning. \textbf{CaMN}~\cite{liu2022beat}, the SOTA model on the BEAT dataset, designs a cascaded architecture and takes into account all six modalities present in the BEAT dataset. All these methods are learned in an end-to-end fashion where the audios are dominated the gesture generation process. We implement their open-source code on two datasets to conduct a fair comparison.

Note that several recent works~\cite{liang2022seeg, ao2022rhythmic} also achieve noticeable performance. We do not compare to SEEG~\cite{liang2022seeg} since they utilize additional data annotations~(Semantic Prompt Gallery). Rhythmic Gesticulator~\cite{ao2022rhythmic} lacks open-source codes. It employs an elaborately designed rhythm-based segmentation strategy to construct its training data, which is different from previous settings~\cite{trimodal, ha2g, liang2022seeg}.

\begin{table}[t]  
\centering
\begin{tabular}{l|ccc}
\hline
Methods      & FGD$\downarrow$ & BC $\uparrow$& Diversity$\uparrow$ \\ \hline
Real Video &  0   &  0.697  & 108.780           \\ \hline
S2G \cite{speech2gesture} & 24.887    & 0.723    &  97.272    \\
TriModal \cite{trimodal} & 4.501    & 0.659    &  102.978    \\
HA2G \cite{ha2g}  &  5.429  & 0.698 &  106.290        \\ \hline
\rowcolor{lightgray!30} Ours Rhythm ($w$=1)  &  2.152  &  0.656  &    107.988    \\ 
Ours Rhythm ($w$=1.5)   &  2.359   &  0.676 &   112.327      \\ 
Ours Rhythm ($w$=2.2)   &  6.622   &  0.699 &    113.051 \\ \hline
Ours Full ($w$=1)   &   11.310  &  0.634 & 108.663    \\ 
Ours Full ($w$=1.5)   &  9.154  & 0.664 & 107.781     \\ 
\rowcolor{lightgray!30} Ours Full ($w$=2.2)   &  8.446   &  0.696 & 109.880   \\ \hline
\end{tabular}
\caption{Comparison with baselines on TED Gesture dataset. Our method outperforms the three baselines in most cases.}
\label{table:tedmetric}
\end{table}

\begin{table}[t]
\centering
\begin{tabular}{l|cccc}
\hline
Methods      & FGD$\downarrow$ & BC $\uparrow$& Diversity$\uparrow$ \\ \hline
Real Video &  0   &  0.867  & 216.541    \\ \hline
S2G \cite{speech2gesture} & 24.887 & 0.872 & 152.367  \\
TriModal \cite{trimodal} & 20.513 & 0.621 &173.214     \\
CaMN \cite{ha2g}  & 8.169  & 0.768 &  183.671       \\ \hline
Ours Rhythm~($w$=1)   &   7.845  &  0.886 & 193.060   \\ 
\rowcolor{lightgray!30} Ours Rhythm~($w$=1.5)   &  7.561  & 0.892 &   206.969       \\ \hline
Ours Full~($w$=1)   & 10.863  &  0.886 &  183.201   \\ 
\rowcolor{lightgray!30} Ours Full~($w$=1.5)   &  9.269  & 0.893 & 194.362   \\ 
\bottomrule
\end{tabular}
\caption{Comparison with baselines on BEAT dataset. Our method achieves the best performance in most cases.}
\label{table:beatmetric}
\end{table}

\subsection{Quantitative Evaluation}

\subsubsection{Rhythm-aware Diffusion Generator} 
\label{sec4:rag}
Thanks to the ability of the diffusion-based method, we can generate the co-speech gestures with varying guidance weights $w$ during inference by employing the classifier-free guidance sampler~\cite{ho2022classifier}. We exhibit the numerical results in the \cref{table:tedmetric} and \cref{table:beatmetric} marked with \textit{Ours Rhythm}. 
We can observe that the Beat Consistency Score and 
Diversity is proportional to the guidance weight. Meanwhile, we achieve our best FGD when $w = 1$. 
As demonstrated in \cref{table:tedmetric}, our RAG is able to beat all baselines in most cases on the TED dataset \cite{trimodal}. S2G~\cite{speech2gesture} achieves the highest Beat Consistency Score. However, our observation reveals that it generates rapid and unnatural body movements regardless of the audio beats, as shown in our supplementary video, resulting in the worst FGD and abnormal Beat Consistency Score, which further shows the weakness of this metric as discussed in recent work~\cite{edge}.
Meanwhile, on the BEAT \cite{liu2022beat} dataset, \textit{Ours Rhythm} and CaMN outperform previous works across all metrics. Despite utilizing only three modalities~(audio, emotion, and speaker ID) as input, our results show comparable FGD as the state-of-the-art model that employs all five modalities. 
Furthermore, our rhythm-aware Diffusion Generator excels in generating diverse and rhythmically complex gestures thanks to our powerful MLP-based diffusion model.

\begin{figure*}
    \centering
    \includegraphics[width=\linewidth ]{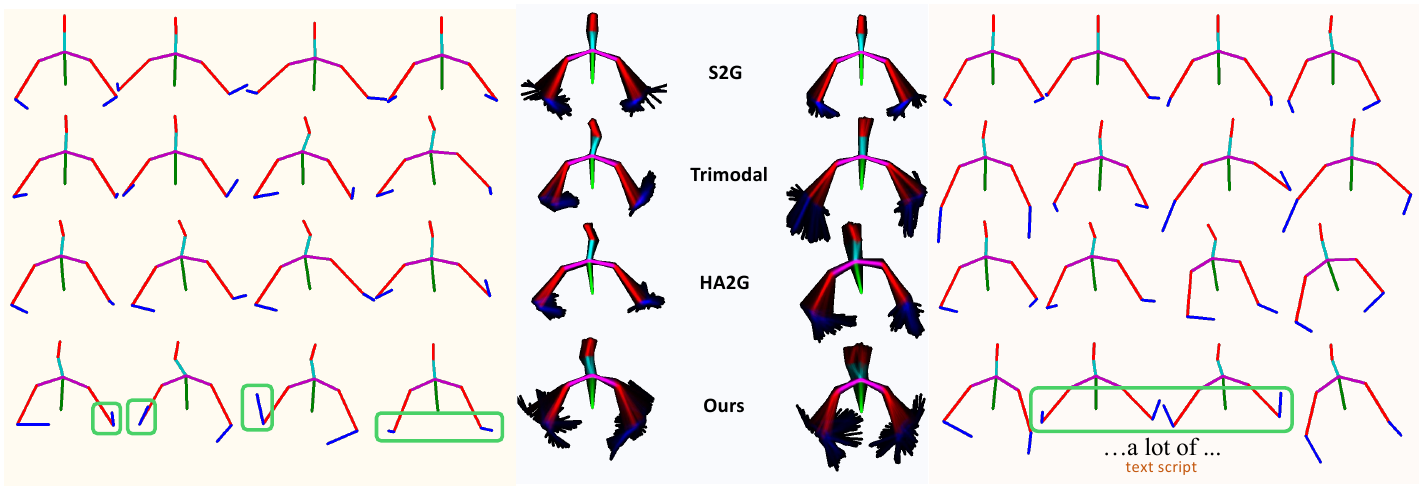}
    \caption{\textbf{Visual comparisons with three baselines}. We display the scope of gestures across frames in the blue background area. Larger scopes mean better diversity. Despite the close Beat Consistency Scores compared to the three baselines, our method stands out by clearly greater diversity as shown in both examples. On the left side of the figure, we roughly show the rhythm of the proposed method. According to \cref{table:tedmetric}, these methods all achieve high and closely comparable BC scores. However, upon visualizing the results, our result excels in terms of diversity. Specifically, our result changes hands from left to right and then waves both hands. In contrast, the baselines maintain a consistent motion pattern throughout. On the right of the figure, we give the text scripts~(``... 
a lot of..."), where the proposed method can produce semantic-aware gestures whether other methods may fail. 
}
    \label{fig:ted_comparison}
\end{figure*}

\subsubsection{Full System}
\label{sec4:full}

As mentioned in \cref{sec3:app}, our RAG has the capability to add beats to any motion sequence in a two-step process of diffusing and denoising through DDIM~\cite{ddim}. We evaluate the whole system as in Fig.~\ref{fig:overview}~(c) with 20 diffuse steps by adding Gaussian noise to the generated motion. The results of our full system are listed in Tab.~\ref{table:tedmetric} and \ref{table:beatmetric} marked with \textit{Ours Full}. On the TED dataset, our full system is also competitive when compared to existing methods. As discussed in SEEG~\cite{liang2022seeg}, the slight downgrade in the FGD can be attributed to the fact that the semantic gestures exhibit worse metrics. A similar observation has been founded in Tab.~\ref{table:beatmetric}, where our full network still keeps the similar Beat Consistency.  

Since human motion is hard to be visualized, we give a simple comparison of the diversity and the semantic-aware gestures in \cref{fig:overview}, where the proposed method generates very diverse gestures than the baseline methods. Besides, the proposed method also shows the semantic-aware gestures from this example. This further shows that the proposed SAG learns some out-of-domain knowledge in terms of FGD and Diversity but still keeps the Beat Consistency. We give more examples in the supplementary video for comparison.

\begin{table}[t]
\centering
\resizebox{\columnwidth}{!}{
\begin{tabular}{l|cccc}
\toprule
Methods    & Natural &  Smooth & Diversity & Semantic  \\ \hline
S2G~\cite{speech2gesture} & 41.6\%   &  36.4\% &  33.3\% & 37.5\%   \\ 
TriModal~\cite{trimodal} & 6.30\%   &  5.20\%  &  10.4\% & 6.30\%  \\
HA2G~\cite{ha2g}  & 9.40\%   &  9.40\% &  10.4\% & 11.4\%   \\ \hline
Ours Full   &  \textbf{42.7\%}   &  \textbf{49.0\%} &  \textbf{45.8\%}   &   \textbf{44.7\%}   \\ \bottomrule
\end{tabular}
}
\vspace{-1em}
\caption{The percentage of the user's favorite methods in terms of naturalness, smoothness, diversity, and semantics.}
\label{table:userstudymetric}
\end{table}

\subsubsection{User Studies}
\label{sec4:user}
Since the generated content is very subjective, we conduct a user study to show the effectiveness of the proposed method over the state-of-the-art methods. Specifically, we ask 16 subjects on four different methods~(\ie, Speech2Gesture~\cite{speech2gesture}, Trimodal~\cite{trimodal}, HA2G~\cite{ha2g} and ours). We provide 12 samples of the results and let them choose the best one in terms of the motion naturalness, the motion smoothness, the diversity of the generated content, and the semantic preservation, yet 768 opinions in total. We then calculate the percentage of each task on each metric. As shown in Tab.~\ref{table:userstudymetric}, the participants like our methods most in terms of four metrics.

\subsection{Ablation Studies}
We ablation two different designs in our method on the TED dataset~\cite{trimodal} for ablation studies. Firstly, we evaluate the effectiveness of the whole system. On the other hand, we ablate the performance of our rhythm-aware diffusion model. More experiments are provided in the supplementary.

\subsubsection{System Overview}
Since each stage is considered individually, we can numerically evaluate the performance of each stage. As shown in ~\cref{table:tedmetric} and Table.~\cref{table:beatmetric}, we have evaluated that our full pipeline can utilize the rhythm of the rhythm-aware generator. Here, we give more ablation studies. As shown in \cref{table:ablation_overall}, although our SAG generates a very different distribution than the original model~(FGD) and bad beat consistency, our single semantic-aware generator gains much more diverse motions than previous methods. As a combination of our two networks~(in the third row), the proposed diffusion model will hugely pull the distribution to the trained one~(as represented by FGD and BC) but still has very diverse results. We also try another method that utilizes the fast Fast Fourier Transform~(FFT) to remove the high-frequency beat information and synthesize the hand-crafted dataset for training the beat alignment network, as shown in the second row of \cref{table:ablation_overall}, this beat alignment network is less effective in terms of FGD and only improve the beat consistency a little.

\subsubsection{Network Structure Ablation on the Rhythm-Aware Diffusion Model}
\label{ab:stage2}
As for the rhythm-aware diffusion model, we set the guidance weights of classifier-free guidance $w=1$ in this section. To validate the effectiveness of our MLP-based model, we replace it with a widely-used transformer structure in the recent motion diffusion model~\cite{tevet2022MDM}. Specifically, we keep others unchanged and utilize the noisy motion sequence as a query and employ the audio features as the key and value to calculate the cross attention. As demonstrated in the second row of \cref{table:ablaiton_rhythm}, our MLP-based model exhibits clear superiority over the model built upon the Transformer Decoder, especially in terms of the Beat Consistency Score. This is crucial for ensuring effective beat empowerment. Meanwhile, we conducted an ablation study on the audio encoder. In this regard, we rebuilt the audio encoder using 2D convolutions and utilized a 128-channel Mel-spectrogram as the audio input. As shown in the third row of Table \ref{table:ablaiton_rhythm}, the result indicates that a relatively simple audio encoder proves to be sufficiently expressive for the task. Additionally, we also evaluate the impact of our loss components. The KL divergence loss item improves FGD and the diversity metrics to a certain extent since it regulates the talking style.

\begin{table}[t]
\centering
\begin{tabular}{l|ccc}
\toprule
Methods   & FGD$\downarrow$ & BC $\uparrow$& Diversity$\uparrow$ \\ \hline
SAG &  56.878  & 0.388   &  128.894        \\ 
SAG + Syn. data & 65.718 & 0.472 & 133.753 \\
LivelySpeaker~(Ours)   & \textbf{8.446}  & \textbf{0.696}  &  109.880  \\
\bottomrule
\end{tabular}
\caption{Ablation Studies on the whole framework where the proposed rhythm-aware Generator can be considered as a stronger beat empowerer.}
\label{table:ablation_overall}
\end{table}

\begin{table}[t]
\centering
\begin{tabular}{l|ccc}
\toprule
Methods   & FGD$\downarrow$ & BC $\uparrow$& Diversity$\uparrow$ \\ \hline
Ours Rhythm~($w=1$)   &  \textbf{2.152}  & \textbf{0.656} &   \textbf{107.988} \\ 
w/ Transformer & 6.509 & 0.418  & 104.737\\
w/ Mel-Spectrogram &4.951 & 0.568 & 101.952\\
w/o KL loss  & 5.256  &  0.650 & 105.126   \\

\bottomrule
\end{tabular}
\caption{Ablation studies on the network structure of the proposed rhythm-aware generator.}
\label{table:ablaiton_rhythm}
\end{table}

\subsection{Limitation}
We propose a diffusion-based rhythm-aware generator that acts as a beat empowerment module, allowing for editing given motion in diffusing first and then denoising manner. Therefore, the inversion steps $K$ are of great significance. For instance, when editing with large diffusing steps $K$~(in extreme cases up to 100), the original motion would be drowned out by the Gaussian noise. If we could get the pair various data of paired sync and out-of-sync data, our results would be further improved via controllable adaptor~\cite{contorlnet}. Similarly, for long sequence generation, individual guidance weight $w$ should also be taken into consideration. As for our semantic-aware generator, its performance is limited by sentence splitting. Take \cref{fig:teaser} as an example, we cannot generate a semantic-aware motion with the bad phrases split, like ‘... from left’ and 'to right...'. Instead of splitting data using a sliding window~(as done in most recent methods), we would pursue a better solution, like pre-parsing the sentence during the training and testing.

\section{Conclusion}
In this paper, we present LivelySpeaker, a novel semantic- and rhythm-aware system for co-speech gesture generation.  To achieve this, we first generate the motion from the semantic-aware generator, after that, we train a diffusion-based rhythm-aware generator and utilize it for rhythm-aware refinement. Powered by our decoupled framework, our method enables multiple new applications in co-speech generation, including text-based pose style controlling, and interpolating between the text- and audio-based gestures. Besides, our pure diffusion-based backbone also achieves state-of-the-art performance in co-speech gesture generation.

\noindent\textbf{Acknowledgments}
The work is supported by NSFC \#61932020, \#62172279, Program of Shanghai Academic Research Leader, and ``Shuguang Program'' supported by Shanghai Education Development Foundation and Shanghai Municipal Education Commission.

{\small
\bibliographystyle{ieee_fullname}
\bibliography{bibliography}
}

\ifarxiv \clearpage In the supplementary material, we provide a \textbf{supplementary video} to show:
\begin{itemize}
    \item The pipeline of the whole model as in the main paper.
    \item The comparison with baselines~(Sec.~\ref{supp:compare}).
    \item The effectiveness of each component in our framework~(Sec.~\ref{supp:ablation}).
    \item The applications of interpolating poses between different modalities~(Sec.~\ref{supp:app1}).
    \item The applications of semantic motion generation via new text prompt~(Sec.~\ref{supp:app2}).
\end{itemize}
We also give some explanations aligned with the video and list below.

\section{Comparisons with baselines}
\label{supp:compare}
We show the results comparing to all baselines~\cite{speech2gesture, trimodal, ha2g, liu2022beat}. On the TED \cite{speech2gesture} dataset, it is noticeable that HA2G \cite{ha2g}, Speech2Gesture~\cite{speech2gesture}, and Trimodal~\cite{trimodal} generate gestures with rhythmic patterns but lack semantic meaning. 
Meanwhile, there exists unnatural arm twitching in HA2G~\cite{ha2g}. 
In contrast, our full pipeline outperforms these baselines by excelling in both semantics and rhythm. 
On the Beat dataset~\cite{liu2022beat}, our method shows better visual performance than the state-of-the-art CaMN~\cite{liu2022beat} that utilizes more modalities. Besides, our approach exhibits greater diversity.

\section{Individual gestures from each generator}
\label{supp:ablation}
We present the generation results of our individual generators. 
As shown in our video, regarding semantics, semantic-aware generator~(SAG) yields open arms for `\textit{many many}', whereas our rhythm-aware generator~(RAG) merely produces waving hands in response to the audio input. However, when the human voice is finished, the output of SAG continues moving while those of RAG become still. Thus, SAG is capable of producing gestures with good content but poor rhythm. 
As for rhythm, our RAG can generate rhythmic-aware results with little semantics.

\section{Application: Interpolating poses between two modalities}
\label{supp:app1}
To combine the merits of SAG and RAG, we employ our RAG as a beat empowerment module, allowing for editing given motion by adding $K$ steps noise first and then, denoising it through a trained gesture diffusion model. By adjusting the value of $K$, we can control the semantics and prosody of gestures as well. Here we exhibit the results under adding different noise steps $K \in \{ 10, 20, 50, 100 \} $. The leftmost one~($K=0$) is the semantic-aware gesture generated from SAG. On its right, we list the edited version of it under different inversion steps. We can observe that when $K$ is small~($\sim 20$), it exhibits both good semantics and rhythm. As the value of $K$ increases, the rhythm-aware gestures dominate the result. However, if the value of $K$ exceeds the threshold~(\eg $K > 50$), the semantic gestures will influence a little.

\section{Application: Semantic gesture generation via new text prompt}
\label{supp:app2}
In our SAG, the motion space is well aligned with the text space of CLIP \cite{radford2021learning}. Inspired by recent advancements in image editing~\cite{p2p, null} through the prompts, we can easily modify and customize the motion in the same manner. As shown in the supplementary video, we present the results directly obtained from SAG, along with the edited outcome achieved by incorporating specific prompts. For instance, we can roughly manipulate the height and range of gestures by providing the prompts such as ``\textit{high}", ``\textit{down}", ``\textit{many}", \etc. We also show an example that when we add the prompt like ``\textit{in a confirm attitude}", it results in a firm waving down motion. We can also observe similar results on the BEAT~\cite{liu2022beat} dataset. Please view our supplementary video for more details. 

\section{Inference speed}
Our two-stage system, which particularly incorporates a diffusion model, is inherently slower during inference time when compared with GAN-based methods.

For speed comparison, we generate a long sequence consisting of 12k frames ($\sim$800s) using each method and report their running time in \cref{table:speed}. The speed of SAG-only is comparable to previous methods while incorporating the diffusion process~($K =$ 20 steps) into our full system increases the running time. Nonetheless, there are various advanced sampling techniques for diffusion models that can be suitable for our method. We believe that future, more advanced sampling techniques can benefit our full pipeline.

\begin{table}[htb]
\resizebox{\linewidth}{!}{
\begin{tabular}{c|ccc|cc}
\hline
 Methods & S2G & TriModal & HA2G & SAG & Ours Full  \\ 
 \hline
Time(s) &  2.9 & 3.1 & 10.8 &  5.4  & 42.6  \\ \hline
\end{tabular}}
\caption{ We conduct the experiment on a single RTX 3090.}
\label{table:speed}
\end{table}

\section{Ablation studies on RAG}
The use of MLPs is inspired by recent work on motion prediction~\cite{du2023agrol}. The 1x1 Conv is a linear layer on the temporal axis. Each MLP block adopts the skip connections, the output from the previous MLP layer is added to the output of its subsequent MLP layer. We choose the hyper-parameter experimentally. Here we present detailed ablation studies on TED in \cref{table:MLP}, where our choice produces the best FGD and Diversity score.

\begin{figure}
    \centering
    \includegraphics[width=\linewidth ]{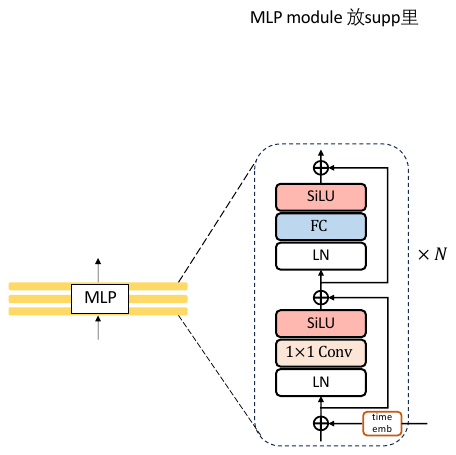}
    \caption{Details of the MLP block.}
    \label{fig:mlp_details}
\end{figure}

\begin{table}[t]  
\centering
\begin{tabular}{cc|ccc}
\hline
        \# &  Act.  & FGD$\downarrow$ & BC $\uparrow$& Diversity$\uparrow$ \\ 
\hline
        4&SiLU   & 2.152 &   0.656  & 107.988   \\  \hline 
        4&ReLU   & 3.956  &  0.683 & 106.581   \\ 
        
        4&LReLU  & 5.847  & 0.682 & 105.668  \\ 
        
        4&LReLU$^\dag$   & 6.392  &  0.695 & 104.497  \\ \hline
        
        2&SiLU   & 8.243  &  0.689 & 106.115 \\ 
        
        6&SiLU   & 3.047 & 0.623 & 104.880    \\ 
        
        8&SiLU  & 4.184  & 0.655 & 104.876 \\ 
                \bottomrule

\end{tabular}
\caption{MLP architecture ablation. LReLU and LRELU$^\dag$ represent the LeakyRELU with the scope of 0.1 and 0.2, respectively. $\#$ represents the layer of MLP in the backbone.}
\label{table:MLP}
\end{table}
 \fi

\end{document}